\pdfoutput=1

\documentclass[11pt]{article}

\usepackage[]{emnlp2021}

\usepackage{times}
\usepackage{latexsym}

\usepackage[T1]{fontenc}

\usepackage[utf8]{inputenc}

\usepackage{microtype}
\usepackage{times}
\usepackage{latexsym}
\usepackage{graphicx}
\usepackage{colortbl}
\usepackage{tabularx}
\usepackage{soul}
\usepackage{arydshln}

\usepackage{titlesec}
\usepackage[utf8]{inputenc}
\usepackage[greek,english]{babel}
\usepackage{alphabeta}
\usepackage{xspace}
\newcommand{\art}{\textsc{art}\xspace}
\newcommand{\es}{\textsc{palo.es}\xspace}
\newcommand{\palo}{\textsc{palo.gr}\xspace}
\newcommand{\semeval}{\textsc{se.en}\xspace}
\newcommand{\semevalplus}{\textsc{se+}\xspace}

\newcommand{\xart}{\textsc{x:art}\xspace}
\newcommand{\xpalo}{\textsc{x:palo}\xspace}
\newcommand{\xartpalo}{\textsc{x:art+palo}\xspace}
\newcommand{\xnope}{\textsc{x:nope}\xspace}
\newcommand{\xzero}{\textsc{x:zero}\xspace}
\newcommand{\gpalo}{\textsc{bert:palo}\xspace}
\newcommand{\rfpalo}{\textsc{rf:palo}\xspace}

\usepackage{times}
\usepackage{latexsym}

\usepackage[T1]{fontenc}

\usepackage[utf8]{inputenc}

\usepackage{microtype}

\title{Analysing the Greek Parliament Records with Emotion Classification}

\author{Vanessa Lislevand \\
  Stockholm University, Sweden \\
  \texttt{lislevand@dsv.su.se} \\\And
  John Pavlopoulos \\
  Ca' Foscari University of Venice, Italy \\
  Stockholm University, Sweden\\
  Athens University of Economics and Business, \\Greece \\
  \texttt{annis@aueb.gr} \\}

\begin{document}
\maketitle
\begin{abstract}
In this project, we tackle emotion classification for the Greek language, presenting and releasing a new dataset in Greek. We fine-tune and assess Transformer-based masked language models that were pre-trained on monolingual and multilingual resources, and we present the results per emotion and by aggregating at the sentiment and subjectivity level. The potential of the presented resources is investigated by detecting and studying the emotion of `disgust' in the Greek Parliament records. We: (a) locate the months with the highest values from 1989 to present, (b) rank the Greek political parties based on the presence of this emotion in their speeches, and (c) study the emotional context shift of words used to stigmatise people.
\end{abstract}

\section{Introduction}
Text emotion detection concerns the classification of a text based on specific emotion categories. The emotion categories are often defined by a psychological model \cite{Oberlander2018} and the field is considered a branch of sentiment analysis \cite{Acheampong2020}. Classifying a text as negative or positive may be a simpler task, but this coarse level of aggregation is not useful in tasks that require a subtle understanding of emotion expression \cite{Demszky2020}. As described by \citet{Seyeditabari2018}, for example, although `fear' and `anger' express a negative sentiment, the former leans towards a pessimistic view (passive) while the latter with a more optimistic one that can lead to action. This has made the detection of emotions preferred over sentiment analysis for a variety of tasks, such as in political science \cite{Ahmad2020}, to measure customer satisfaction in marketing \cite{bagozzi1999}, to use the emotional state of the user in recommendation systems \cite{brave2002}, and to monitor the public sentiment during crisis \cite{Kabir2021}.

\begin{figure}[t]
\centering
  \includegraphics[width=.45\textwidth]{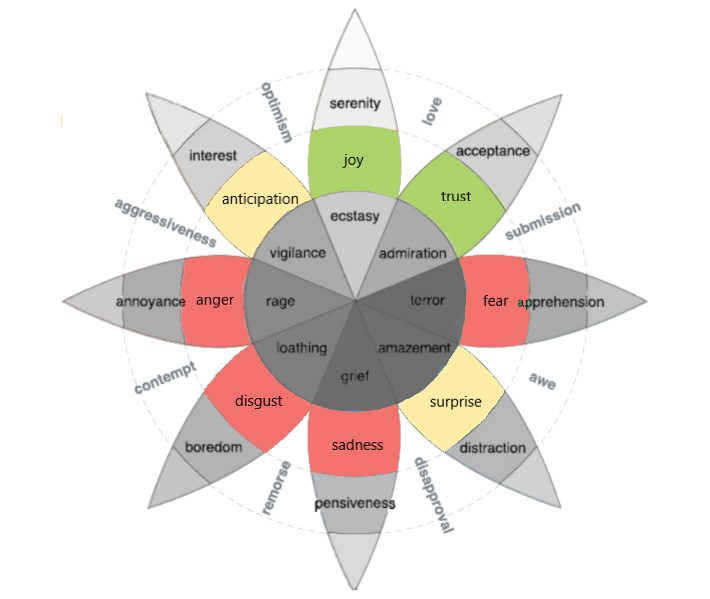}  \caption{\footnotesize Plutchik’s Wheel of emotions colored based on our sentiment aggregation. Green colour corresponds to positive sentiment, red to negative sentiment, and yellow to emotions that we didn't include in the aggregation.}
  \label{fig:wheel}
\end{figure}

Most of the work in emotion detection concerns resource-rich languages and only few published studies concern emotion detection for under-represented languages \cite{Ahmad2020}. The same problem exists for the Greek language, for which there is no publicly available dataset for the task. This study focuses on the detection of the eight basic emotions from Plutchik’s Wheel \cite{plutchik}, shown in Figure~\ref{fig:wheel}, for the Greek language. We developed a new
Greek dataset which we release for public use. \footnote{\url{https://github.com/ipavlopoulos/emotion.gr}} We used this dataset to benchmark a multilingual and a monolingual Transformer-based emotion classifier, and to assess them for emotion classification. Our results show that although our benchmarks achieve low to average results for most of the studied emotions, the performance for one of the emotions, vis. \textsc{disgust}, is much higher and comparable to the performance of sentiment and subjectivity classification when we aggregate the emotions accordingly. 

By employing an emotion classifier trained on our data, we analysed the records of the Greek Parliament Corpus from 1989 to 2020 with regards to \textsc{disgust}, the most frequently occurring emotion in electoral data \cite{Mohammad2015}. First, we studied the points in time when this emotion was most frequently occurring in the Greek parliament records and we ranked the Greek political parties based on the detected score. Second, we used it to investigate the emotional context shift, focusing on words used to stigmatise. Our findings show that the records of the far right wing are relatively higher in \textsc{disgust}, when compared to the rest, and (b) words that could stigmatise people are being increasingly used in an emotional context related to \textsc{disgust} in the studied parliamentary records. 

\section{Related work}
Emotion classification is a natural language processing (NLP) problem with various use cases \cite{Oberlander2018,Acheampong2020,Demszky2020,Seyeditabari2018,sailunaz2018,Gaind2019}.\footnote{An older review of the field can by found in the work of \citet{Mohammad2016}.} Following the introduction of Transformers \cite{vaswani2017}, several Transformer-based monolingual and multilingual pre-trained language models \cite{Devlin2018,conneau} (PLMs) have been on the spotlight, leading to improvements for many NLP tasks through their fine-tuning,  including emotion and sentiment classification. \citet{kant2018} demonstrated that pre-training an attention-based Transformer network on Amazon reviews and then fine-tuning it on the \textit{‘SemEval Task 1: E-c’} multilabel emotion classification training set offers benefits \cite{mohammad2018}. We also explore the benefits of using this dataset, as we will discuss later.
\citet{desai2020} presented HURRICANE EMO, a dataset with 15,000 tweets, each annotated with emotions perceived after hurricanes. They suggested classification tasks to discriminate between coarse grained emotion groups and experimented with neural networks and PLMs.

\subsubsection*{Emotion detection in not well resourced settings}

The above mentioned advancements have assisted research more broadly in NLP for high-resource languages such as English, but the challenge remains for the rest.
\citet{ranasinghe2020} experimented with transfer learning from English to Bengali, Hindi and Spanish at the offensive language identification task. They showed the superiority of XLM-R \cite{conneau}, a multilingual BERT-based PLM, compared to the best systems submitted to recent shared tasks in these three languages. \citet{Ahmad2020} implemented emotion detection in Hindi proposing a deep transfer learning framework from English which captures relevant information through the shared embedding space of the two languages.

\citet{tela2020} compared the English XLNet \cite{xlnet}, fine-tuned with the Tigrinya language with a new monolingual Transformer (TigXLNet), pre-trained on Tigrinya. Even though the XLNet was trained with only 10k examples of the Tigrinya sentiment analysis dataset, its results were comparable to the results of TigXLNet and it outperformed both BERT \cite{Devlin2018} and mBERT \cite{Devlin2018}. \citet{hedderich2020} evaluated transfer learning and distant supervision on mBERT and XLM-R, from English to three African languages Hausa, isiXhosa, and Yoruba, on named entity recognition and topic classification. They showed that even with a small amount of labeled data, a reasonable performance can be achieved. \citet{pires2019} studied the performance of mBERT at zero-shot cross-lingual model transfer, by fine-tuning the model using task-specific supervised training data from one language, and evaluating on that task in a different language. They showed that transfer is possible even to languages in different scripts, and that better performance is achieved when the languages are typologically similar. 

\citet{lauscher2020} studied the effectiveness of cross-lingual transfer for distant languages through multilingual transformers. They detected a correlation between the performance of transfer-learning with: (1) the philological similarity between the source and the target languages, and (2) the size of the corpora used for pre-training on target languages, which highlighted the significance of few-shot transfer-learning. \citet{das2021} showed that XLM-R outperformed various machine and deep learning, and Transformer-based approaches at the emotion classification task for the six basic emotions (i.e., \textsc{anger, fear, disgust, sadness, joy}) and for \textsc{surprise} for the Bengali language. \citet{kumar2021} evaluated XLM-R with zero-shot transfer learning from English to Indian on the Benchmark `SemEval 2017 dataset Task 4 A' \citet{semeval2017}, and proved that the model compares favorably to state-of-the-art approaches.

\subsubsection*{Emotion detection for the Greek language}
Although there are a few published studies focusing on sentiment analysis in Greek \cite{markopoulos2015,athanasiou2017}, limited published work concerns emotion detection, probably due to the lack of publicly available resources. Fortunate exceptions are the work of \cite{Krommyda2020} and the work of \cite{Palogiannidi2016}. The authors of the former study suggested the use of emoji in order to assign emotions to a text, but they didn't share their dataset while this approach is expected to work only with emoji-rich corpora. We also experimented with a dataset created with indicators of emotions, which we release for public use. The latter study created an affective lexicon, which can lead to efficient solutions, but not useful to train machine and deep learning algorithms, such as BERT \cite{koutsikakis2020} that achieve the state of the art nowadays. \citet{Alexandridis2021} was the first to experiment with two BERT-based models, trained on a Greek emotion dataset. This dataset can be used to train supervised learning solutions, but unfortunately it is not publicly available. Upon communication with one of the authors, part of their data are comprised in our dataset. To the best of the authors' knowledge, this is the first work focusing on emotion detection for the Greek language, employing (monolingual and multilingual) deep learning solutions, while releasing the developed datasets to promote future research.

\section{Dataset development}
This section presents our new dataset, comprising tweets annotated regarding the emotion of the author. We discuss its three parts, the evaluation (\es), the training (\palo), and the one used for augmentation (\art). Also, we discuss how we extended a well-known English emotion detection dataset, used to fine-tune PLMs first in English, with neutral tweets, in order to adjust to a setting where the majority of tweets is characterised by lack of emotion. 

\noindent \textbf{The artificial \art dataset} was developed by retrieving Greek tweets for several emotions using Tweepy.\footnote{\url{https://www.tweepy.org/}} In order to retrieve tweets per emotion, we used target words that could have been selected by users under specific emotional states. For example, in order to collect tweets related to \textsc{joy}, we searched for tweets that contain words such as \textit{‘χαίρομαι’} (\textit{‘I am happy’}). The exact words that were used to retrieve tweets for each emotion are presented in Table~\ref{table:art}.\footnote{We note that not all words referring to a specific emotion lead to the retrieval of tweets comprising that emotion. For example, searching for ‘χαρά’ (‘happiness’; aiming for tweets classified to \textsc{joy}), we receive emotionless tweets, such as `Η χαρά είναι ένα συναίσθημα που πρέπει να εκφράζεται στον ίδιο βαθμό όπως και τα υπόλοιπα' (`Happiness is an emotion that must be expressed to the same degree as the rest.').}

\begin{table}[]
\centering
\scalebox{0.5}{
\begin{tabular}{|l|l|} 
\hline
Words & Category \\ \hline\hline
\begin{tabular}[c]{@{}l@{}}‘αίσχος’(disgrace), ‘έλεος’(mercy), ‘ήμαρτον’(drat), ‘αι σιχτιρ’(get lost?),\\ ‘γαμώ’(fuck),‘νιώθω εξοργισμένος’(I feel angry), ‘νιώθω οργή’\\(Ι feel anger), ‘βλάκα’(fool),‘ηλίθιος’(stupid), ‘σιχαμα’(abomination)\end{tabular} & anger,~disgust  \\ 
\hline
‘περιμένω’(wait), ‘αναμένω’(expect), ‘προσμένω’(look forward) & anticipation \\ 
\hline
\begin{tabular}[c]{@{}l@{}}
‘φοβάμαι’(Ι am afraid), ‘τρομάζω’(scare), ‘τρομακτικό’(scary),\\ ‘τρέμω’(tremble), ‘σκιάζομαι’(?)\end{tabular} & fear \\ 
\hline
\begin{tabular}[c]{@{}l@{}}‘χαίρομαι’(Ι am glad), ‘είμαι χαρούμενος’(I am happy), ‘πολύ χάρηκα’\\(I was very happy), ‘αχ ναιιι’(oh yeahhh),‘ναιιι’(yesss), ‘τέλειοο’(perfect),\\ ‘εκστασιασμένος’(ecstatic)\end{tabular}                    & joy             \\ 
\hline
\begin{tabular}[c]{@{}l@{}}
‘λυπάμαι’(I am sorry), ‘στεναχωριέμαι’(feel sad), ‘θλίβομαι’(grieve),\\ ‘θλίψη’(sadness), ‘απογοήτευση’(dissapointment) \end{tabular}                                                                                             & sadness         \\ 
\hline
‘εκπλήσσομαι’(I am surprised, ‘έκπληξη’(surprise)                                                                                                                                     & surprise        \\ 
\hline
‘εμπιστοσύνη’(trust), ‘εμπιστεύομαι’(I trust)                                                                                                                                & trust           \\
\hline
‘ανακοίνωση’(announcement), ‘είδηση’(news)                                                                                                                                & none           \\
\hline
\end{tabular}
}
\caption{Words used to retrieve tweets per emotion for the development of \art.}
\label{table:art}
\end{table}

\begin{table}[]
    \centering
    \scalebox{0.85}{
    \begin{tabular}{|c|l|}\hline
         Class & Emotions \\\hline\hline
         \textsc{anger} & anger, annoyance, rage \\\hline
         \textsc{anticipation} & anticipation, interest, vigilance \\\hline
         \textsc{disgust} & disgust, disinterest, dislike, loathing\\\hline
         \textsc{fear} & fear, apprehension, anxiety, terror\\\hline
         \textsc{joy} & joy, serenity, ecstasy\\\hline
         \textsc{sadness} & sadness, pensiveness, grief\\\hline
         \textsc{surprise} & surprise, distraction, amazement\\\hline
         \textsc{trust} & trust, acceptance, liking, admiration\\\hline
         \textsc{other} & sarcasm, irony, or other emotion\\\hline
         \textsc{none} & no emotion\\\hline
    \end{tabular}
    }
    \caption{Emotion classes and their respective emotions.}
    \label{tab:emotions}
\end{table}

\begin{table*}
\centering
\small{
\begin{tabular}{l c c c c c c c c c |c } 
& \sc anger & \sc antic. & \sc disgust       & \sc fear  & \sc joy  & \sc sadness & \sc surprise & \sc trust & \sc none & \sc TOTAL\\ 
\hline
\semeval  & \textbf{37.0} & \textbf{14.3} & \textbf{37.8} & \textbf{17.6} & \textbf{37.2} & \textbf{29.4} & 5.1 & 5.2 & 2.8 & 7,724\\\hline
\semevalplus  & 33.6 & 12.9 & 34.3 & 16.0 & 33.8 & 26.7 & 4.6 & 4.7 & 11.9 & 8,519  \\ \hline
\art  & 12.9 & \textbf{12.9} & 12.9 & 12.9 & 12.9 & 12.9  & \textbf{10.9} & 11.7 & 12.9  & 7,753\\  \hline
\palo  & 9.8 & 9.8 & 24.2 & 0.7 & 16.2 & 1.5 & 6.2 & \textbf{21.6} & 46.2  & 2,408\\ \hline
\es & 10.8 & 2.8 & 31.7 & 0.5 & 1.8 & 0.6 & 1.4 & 2.2 & \textbf{60.6} & 786 \\
\end{tabular}
}
\caption{The relative frequency per emotion (columns 1-8), or their absence (column 9), along with the total number of tweets (last column) per dataset. In bold are the highest values per class.}
\label{table:data support}
\end{table*}

\noindent \textbf{The evaluation \es dataset} comprises Greek tweets provided by
\textit{Palo Services},\footnote{\url{http://www.paloservices.com/}} each annotated by two professional annotators employed by the company. Each tweet was annotated regarding ten emotion classes, which are presented in Table~\ref{tab:emotions}. As a first step, we shared a small sample of one hundred tweets to estimate inter-annotator agreement, providing no specific instructions. Cohen’s Kappa was found to be as low as 0.29. In a second annotation round, we provided the annotators with the guidelines suggested by \citet{mohammad2018} and asked two questions per tweet. Our first question was: \textit{Which of the following options best describes the emotional state of the tweeter?}, seeking for the primary emotion of the respective tweet. The second question was: \textit{Which of the following options further describes the emotional state of the tweeter? Select all that apply.}, now allowing more than one emotions to be assigned. Tweets were provided to the annotators as examples per emotion (more details can be found in Table~\ref{table:examples} of Appendix~\ref{appendix:data}). Cohen's Kappa improved to 0.36 for the primary emotions while Fleiss Kappa \cite{fleiss} was found to be 0.26 for the multi-label annotation setting. 

A manual investigation of the annotations revealed that disagreement was often on tweets comprising news or announcements. Attempting to alleviate a possible misunderstanding, we updated the annotation guidelines so that the annotators were guided to classify tweets with news or announcements to the \textsc{none} class (more details can be found in Table~\ref{tab:ext_guideline} of Appendix~\ref{appendix:data}). With the updated guideline we proceeded to the final annotation round by providing both annotators with the same batch of 999 tweets and filtering out tweets that the annotators disagreed on. Cohen's Kappa improved to 0.51 (+15) and Fleiss Kappa improved to 0.44 (+18). We kept 786 out of 999 tweets that annotators agreed on at least one emotion, rejecting 146 tweets with no agreement and 68 tweets labeled with the emotion \textsc{other}. Due to its size and guaranteed quality, we employ \es only for evaluation purposes.

\noindent \textbf{The training \palo dataset} follows the same annotation process as with \es, but each professional annotator was now given 1,000 different tweets. Out of the 2,000 annotated tweets, we excluded 135 (6.8\%) that were labelled as \textsc{other}, leaving 1,865 tweets in total. In order to augment the under-represented positive emotion classes (i.e., \textsc{anticipation, joy, surprise, trust}), we  provided our annotators with 543 more tweets, which had been classified as positive with prior annotation efforts by the company. This led to a total of 2,408 tweets.

\noindent \textbf{Using an existing English dataset} can assist as a prior step, by fine-tuning multilingual PLMs in emotion detection in English, before moving to a resource-lean language, such as Greek.
\citet{mohammad2018} introduced such a dataset for the 1st SemEval E-c Task, a multi-dimensional emotion detection dataset,\footnote{\url{https://competitions.codalab.org/competitions/17751}} which can be used to fine-tune (multilingual or monolingual) PLMs in emotion classification in English. We will refer to this dataset as \semeval. The task of the challenge was defined as: \textit{“Given a tweet, classify it as ‘neutral or no emotion’ or as one, or more, of eleven given emotions that best represent the mental state of the tweeter”}. The dataset consisted of 7,724 tweets with binary labels for each of the eight categories of \citet{plutchik}: \textsc{anger, fear, sadness, disgust, surprise, anticipation, trust}, and \textsc{joy}, which were expanded with \textsc{optimism, pessimism, love}, and with \textsc{none} for the neutral tweets. These categories are not mutually exclusive, i.e., a tweet may belong to one or more categories. For example, the tweet: \textit{‘Don’t be afraid to start. Be afraid not to start. \#happiness’}, belongs to two classes: \textsc{fear} and \textsc{joy}.

Neutral \semeval tweets (in train and dev sets) were 218 (2.8\%), which means that it is assumed that most often tweets do comprise emotions. Although this may be simply due to the sampling of the data, we find that this assumption is weak. Depending on the domain, most often it is the lack of emotion that characterises a tweet. Based on this observation, and in order to better represent the neutral class, we enriched \semeval with 795 neutral tweets that were taken from the timeline of the British newspaper \textit{‘The Telegraph’},\footnote{\url{https://www.telegraph.co.uk/}} and provided by the online community Kaggle.\footnote{\url{https://www.kaggle.com/}} We will refer to this extended English dataset as \semevalplus.\footnote{Preliminary experiments with the dataset of \citet{Demszky2020} showed that it wasn't beneficial.}

\noindent \textbf{The class support of all the datasets} is presented in Table~\ref{table:data support}. \semevalplus has the highest total support and the highest percentage of the categories \textsc{anger, anticipation, disgust, fear, joy} and \textsc{sadness} compared to the other datasets. The distribution of the support for the \art dataset is spread. For the \palo and \es datasets a high percentage for the category \textsc{disgust} and especially for the category \textsc{none}. By adding more neutral tweets to \semeval, the support for \textsc{none} increased from 2.8\% to 11.9\%, almost reaching that of \art (12.9\%). 

\section{Empirical analysis}
We preprocessed the tweets of all the datasets by removing all URLs and usernames (e.g., \textit{@Papadopoulos}) while tokenisation was undertaken with respect to each model's properties. We trained our systems in order to classify the tweet into one or more of the eight former emotion categories of Table~\ref{table:emotion}, excluding \textsc{none}. The score for the \textsc{none} class was calculated as the complementary of the maximum probability of the other eight categories. In other words, if the maximum emotion score was lower than 0.5, the NONE class was assigned.

\subsubsection*{From emotions to subjectivity and sentiment}
In order to study not only the emotions but also the sentiment of the tweets, we aggregated \textsc{anger, fear, sadness, disgust} into a \textsc{`negative'} sentiment category (in red in Fig.~\ref{fig:wheel}). \textsc{trust} and \textsc{joy} were aggregated into a \textsc{`positive'} category (in green in Fig.~\ref{fig:wheel}). The rest were considered as belonging to a \textsc{`neutral'} category. \textsc{anticipation} and \textsc{surprise} were not considered neither as \textsc{positive} nor as \textsc{negative}, because we find that the sentiment that they express is ambiguous. To model subjectivity, we used the \textsc{none} emotion class, linking low \textsc{none} scores to the subjective class and high scores to the objective.

\subsubsection*{Opted evaluation measure}
For evaluation, we report the Area Under Precision-Recall Curves (AUPRC) per emotion, sentiment and subjectivity category, which was preferred due to the highly imbalanced nature of our dataset.\footnote{AUPR captures the tradeoff between precision and recall for different thresholds.}

\subsection{Machine and deep learning benchmarks}\label{Benchmarks}

\begin{table*}[!ht]
\centering\small
\begin{tabular}{l c c c c c c c c c |c }
& \textsc{anger} & \textsc{antic.} & \textsc{disgust} & \textsc{fear} & \textsc{joy}  & \textsc{sadness} & \textsc{surprise} & \textsc{trust} & \textsc{none} & \textsc{AVG}  \\ \hline
\xzero  & 0.38 & 0.12 & 0.82 & 0.03 & 0.49 & 0.10 & 0.07 & 0.18 & 0.92 & 0.35 \\ \hline
\xart & 0.33 & 0.13 & 0.68 & 0.07 & 0.31 & 0.07 & 0.05 & 0.10 & 0.89 & 0.29 \\ \hline
\xartpalo & \textbf{0.51} & 0.43 & 0.94 & \textbf{0.15} & 0.50 & \textbf{0.19} & 0.06 & 0.25 & \textbf{0.99} & \textbf{0.45} \\ \hline
\xpalo & 0.46 & \textbf{0.50} & 0.93 & 0.09 & \textbf{0.54} & 0.04  & \textbf{0.09} & \textbf{0.28}  & \textbf{0.99} & 0.44 \\\hline
\xnope & 0.43 & 0.19 & 0.90 & 0.03 & 0.48 & 0.03 & 0.03 & 0.20 & 0.98 & 0.37 \\ \hline
\gpalo & 0.49 & 0.31 & \textbf{0.95} & 0.03 & 0.45 & 0.03 & 0.03 & 0.24 & 0.98 & 0.39 \\ \hline
\rfpalo & 0.34 & 0.14 & 0.81 & 0.05 & 0.13 & 0.02 & 0.03 & 0.10 & 0.93 & 0.28 \\
\end{tabular}

\caption{Emotion classification AUPRC per emotion and macro-average across all emotions (last column). The average across three restarts is shown per model.}
\label{table:emotion}
\end{table*}

\begin{table}[!ht]
\centering
\arrayrulecolor{black}
\scalebox{0.7}{
\begin{tabular}{!{\color{black}\vrule}l!{\color{black}\vrule}l!{\color{black}\vrule}l!{\color{black}\vrule}l!{\color{black}\vrule}!{\color{black}\vrule}l!{\color{black}\vrule}!{\color{black}\vrule}l!{\color{black}\vrule}l!{\color{black}\vrule}!{\color{black}\vrule}l!{\color{black}\vrule}} 
\hline
                      & \multicolumn{3}{l!{\color{black}\vrule}!{\color{black}\vrule}}{\textbf{Sentiment}} &               & \multicolumn{2}{l!{\color{black}\vrule}!{\color{black}\vrule}}{\textbf{Subjectivity}} &                \\ 
\hline
               & \textbf{neg}  & \textbf{pos}  & \textbf{neu}                                       & \textbf{AVG} & \textbf{subj}  & \textbf{obj}                                                          & \textbf{AVG}  \\ 
\cline{1-7}\arrayrulecolor{black}\cline{8-8}
\xzero            & 0.84          & 0.40          & 0.93                                               & 0.72          & 0.80          & 0.93                                                                  & 0.86           \\ 
\arrayrulecolor{black}\hline
\xart       & 0.69          & 0.18          & 0.90                                               & 0.59          & 0.72          & 0.90                                                                  & 0.81           \\ 
\hline
\xartpalo & 0.95          & 0.41          & \textbf{0.99}                                      & 0.78          & \textbf{0.97} & \textbf{0.99}                                                         & \textbf{0.98}  \\ 
\hline
\xpalo      & 0.95          & \textbf{0.43} & \textbf{0.99}                                      & \textbf{0.79} & 0.96          & \textbf{0.99}                                                         & \textbf{0.98}  \\ 
\hline
\xnope           & 0.93          & 0.39          & \textbf{0.99}                                      & 0.77          & 0.95          & \textbf{0.99}                                                         & 0.97           \\ 
\hline
\gpalo         & \textbf{0.96} & 0.39          & \textbf{0.99}                                      & 0.78          & \textbf{0.97} & \textbf{0.99}                                                         & \textbf{0.98}  \\ 
\hline
\rfpalo               & 0.84          & 0.17          & 0.95                                               & 0.65          & 0.87          & 0.95                                                                  & 0.91           \\
\hline
\end{tabular}
}
\arrayrulecolor{black}
\caption{AUPRC in sentiment and subjectivity classification, using our eight emotion classifiers (the average across three restarts is shown). The two macro average scores are shown on the right of each task.}
\label{table:sent and subj}
\end{table}

\begin{figure}[t]
\centering
  \includegraphics[width=.5\textwidth]{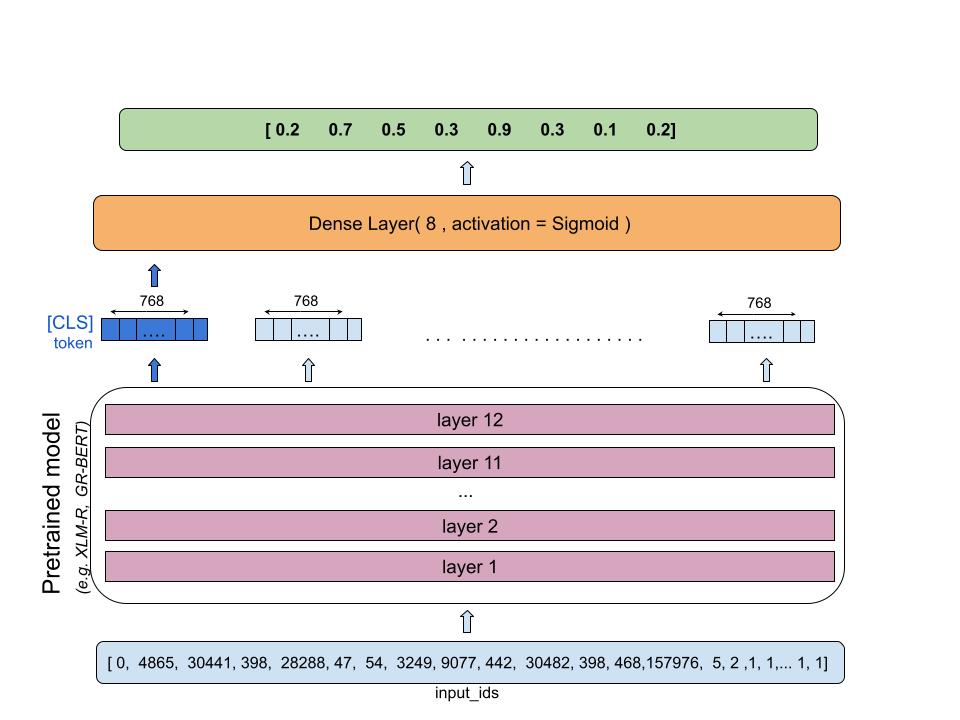}  \caption{\footnotesize The architecture of XLM-R and GreekBERT for the emotion classification task.}
  \label{fig:model}
\end{figure}

We used six Transformer-based models, using one PLM pre-trained on multiple languages and one that was pre-trained on Greek. As a simpler baseline, we opted for Random Forests (\rfpalo).\footnote{We used TFIDF and default parameters of: \url{https://scikit-learn.org/stable/}.}

\noindent \textbf{XLM-R} \cite{conneau} is a Transformer-based multilingual PLM which leads to state-of-the-art performance on several NLP tasks, especially for low-resource languages. For our task, we added a fully-connected layer on top of the pre-trained XLM-R model. We fed the pre-trained model with vectors that represent the tokenised sentences, and subsequently, the pre-trained model fed the dense layer with its output, i.e., the context-aware embedding (length of 768) of the [CLS] token of each sentence. The number of nodes in the output layer is the same as the number of classes (eight). Figure~\ref{fig:model} illustrates the architecture of the system. We fine-tuned the multilingual XML-R first on the English \semevalplus and then we further fine-tuned it on the Greek \art and \palo datasets, yielding two models: \xart and \xpalo respectively. We also experimented with merged \art and \palo, yielding \xartpalo. To assess the benefits of using an English dataset as a prior step, we fine-tuned XLM-R directly on \palo, without any fine-tuning on \semevalplus, which yielded \xnope. and we also tried zero-shot learning by training the model only on \semevalplus, yielding to \xzero.

\textbf{GreekBERT} was introduced by \citet{koutsikakis2020} and it is a monolingual Transformer-based PLM for the modern Greek language. The architecture of the model is similar with XLM-R, as can be seen in Figure~\ref{fig:model}.\footnote{We used: \url{https://huggingface.co/}.} We fine-tuned GreekBERT on \palo, which led to the \gpalo model. Experimental details for all our models are shared in Appendix~\ref{appendix:exp}.

\subsection{Experimental Results}
We used as our evaluation set the high quality \es dataset and we present the results in emotion, sentiment and subjectivity classification. 

\subsubsection*{Emotion classification}
Table~\ref{table:emotion} presents the AUPRC (average across three restarts) of all eight models, per class and overall, for the task of emotion classification. The standard error of mean is also calculated and shared in Appendix~\ref{appendix:exp} (Table~\ref{table:emotion-sem}). \xartpalo was the best overall, achieving the best performance in \textsc{anger, fear, sadness} and \textsc{none}. \xpalo followed closely, with best performance in \textsc{anticipation, joy, surprise, trust} and (shared) in \textsc{none}.

\subsubsection*{Sentiment and subjectivity classification}
Table~\ref{table:sent and subj} presents the AUPRC for the task of sentiment and subjectivity detection. \xartpalo, \xpalo and \gpalo perform equally high in subjectivity (0.98). These models were also top performing for the neutral sentiment and the objective class, along with the \xnope model, which did not use fine-tuning in English as a prior step. This means that using an English dataset as a prior fine-tuning step, assisted in the detection of the subjective emotions. In specific, \xpalo was the best for positive and \gpalo for negative ones.    

\subsubsection*{Zero-shot classification}
Considering its zero-shot learning, \xzero achieved considerably high scores in \textsc{disgust} and \textsc{none} (0.82 and 0.92 respectively), also scoring high in \textsc{joy}. More generally for \textsc{positive} emotions, it scored only three units lower from the best performing \xpalo. \xzero also outperformed \xart, which had the worst results. The low performance  of \xart indicates that retrieving data based on keywords may not be the right way to build a training dataset, when the evaluation dataset is sampled otherwise. On the other hand, combined with other datasets it can lead to improvements, as for example \xartpalo that outperforms both \xart and \xpalo for the emotion classification task, and especially for subjective emotions.

\subsubsection*{Emotion classification averaged across systems}

\begin{figure}[ht]
\centering
  \includegraphics[width=.45\textwidth]{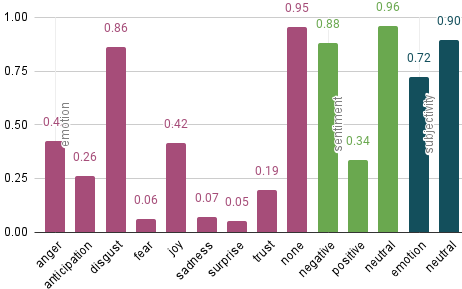}  \caption{Average AUPRC score of all eight systems in emotion (in purple), sentiment (light green), subjectivity (dark green) classification.}\vspace{-5pt}
  \label{fig:avg_sys}
\end{figure}

Figure~\ref{fig:avg_sys} presents the average (across systems) AUPRC score per emotion, sentiment and subjectivity class. This organisation of the results allows us to compare the different emotions and emotion groups for the average system. We observe that our dataset provide adequate training material for \textsc{disgust} and for the lack of any emotion (\textsc{none}). The former probably explains also the high score for the \textsc{negative} sentiment while the latter probably explains the high score for the \textsc{neutral} class.

\section{Detecting emotions in political speech}
We mechanically annotated and studied the emotion in the textual records of the Greek Parliament. We focused on the \textsc{disgust} emotion, which is the emotion that our classifiers capture best (see Figure~\ref{fig:avg_sys}). We opted for detecting a single emotion, instead of sentiment or subjectivity, because the latter could be linked to multiple emotions and hence provide us with an inaccurate signal. For example, `fear' and `anger' are both negative, but the pessimistic view of the former differs from the optimistic view of the latter \cite{Seyeditabari2018}. Such subtle differences, however, should not be ignored in our socio-political study \cite{Ahmad2020}, where we: (a) explore the emotion evolution in political speech, (b) utilise its presence to compare Greek political parties, (c) explore the context of terms used to stigmatise people \cite{Rose2007}. 

\noindent\textbf{The Greek Parliament Corpus},\footnote{\url{https://doi.org/10.5281/zenodo.2587904}} which we used to undertake this study, comprises 1,280.918 speeches of Greek Parliament members from 1989 to 2020, which were split into 9,096.021 sentences (with average word length of 19) for the purposes of our research.
Preliminary experiments with our top three emotion classifiers, \xpalo, \gpalo, \xartpalo, showed that the first performs best. Hence, we used \xpalo to perform the emotion/sentiment analysis that we discuss next.

\begin{figure}[t]
\centering
  \includegraphics[width=.45\textwidth]{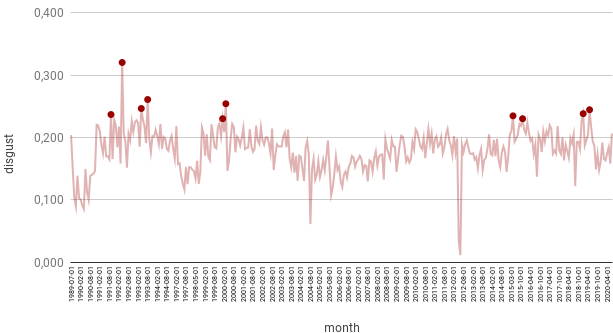}  \caption{\footnotesize Average predicted \textsc{disgust} score per month for the records of the Greek Parliament Corpus. The ten highest values are shown with red bullets.}\vspace{-10pt}
  \label{fig:time}
\end{figure}

\subsection{`Disgust' evolution in political speech}\label{ssec:political_disgust} Figure~\ref{fig:time} illustrates the detected \textsc{disgust} emotion, monthly averaged, with the highest values (i.e., months) highlighted. A probability score was computed for each sentence of the Greek Parliament records, by employing the \textsc{disgust} emotion head of our \xpalo model (the model selection process is discussed in Appendix~\ref{appendix:emo_in_pol}). Then, we macro-averaged the computed scores per month. The highest \textsc{disgust} score was observed between 1991 and 1993 (i.e., September 1991, April 1992, April 1993, August 1993) in 2000 (i.e., January 2000, March 2000), in 2015 (i.e., November 2015, April 2015) and in 2019 (i.e., January 2019, May 2019). By investigating the main events of each of these months, we find that there is at least one event per month that could potentially rationalize these high scores (more information about the selected events and some examples of text may be found in Table~\ref{table:events} and Table~\ref{table:examples-disgust} in Appendix~\ref{appendix:emo_in_pol}).

\subsection{Greek political parties related to `disgust'}
By computing the average \textsc{disgust} score per party,\footnote{We computed one score per sentence and macro-averaged across all the sentences of the respective party.} we were able to compare all political parties, as depicted in Figure~\ref{fig:political}. We observe that the score is relatively high for the top two rows, which correspond to the far right wing. The \textit{Democratic Social Movement} and the \textit{Communist Party of Greece} follow closely. On the lower end of the diagram are the \textit{Opposition} and the \textit{Parliament}. \textit{Opposition} consists of one or more political parties that are opposed, primarily ideologically, to the government, party or group in political control of the country. \textit{Parliament} is simply the coordinator of the speakers in the Parliament. Both these two are characterised by lack of any emotion, which can be explained by the template sentences that they use in their speeches. For example, the most common sentence of the \textit{Parliament} is `Μάλιστα, μάλιστα' (translated as: \textit{`Affirmative, affirmative'}). Correspondingly, a common sentence of \textit{Opposition}is the `Κατά πλειοψηφία' (translated as: \textit{`By majority.'}). However, the \textsc{disgust} of \textit{Opposition} is higher than that of \textit{Parliament}, most probably because the former also comprises sentences that could express \textsc{disgust}, such as: 'Αίσχος, αίσχος' (translated as: \textit{`Disgrace, disgrace'})

\subsection{Evolution of `disgust' for target terms}
\begin{table}\small
\begin{tabular}{|l|c|} \hline
  \bf Political Party & \bf Score \\\hline
  \cellcolor{red!33}Golden Dawn & 33\% \\
  \cellcolor{red!28.6}Greek Solution & 28.6\%\\
  \cellcolor{red!28.3}Democratic Social Movement & 28.3\%\\
  \cellcolor{red!26.4}Communist Party of Greece & 26.4\%\\
  \cellcolor{red!25.2}Alternative Ecologists & 25.2\%\\
  \cellcolor{red!24.6}Political Spring & 24.6\%\\
  \cellcolor{red!24.5}Independent (out of party) & 24.5\%\\
  \cellcolor{red!23.8}Independent Democratic MPs & 23.8\%\\
  \cellcolor{red!23.5}Center union & 23.5\%\\
  \cellcolor{red!21.6}Democratic Alliance & 21.6\%\\
  \cellcolor{red!21.5}Coalition of the Radical Left & 21.5\%\\
  \cellcolor{red!20.7}Coalition of the Left, of Movements and Ecology & 20.7\%\\
  \cellcolor{red!20.7}European Realist Disobedience Front & 20.7\%\\
  \cellcolor{red!20.6}Independent Greeks & 20.6\%\\
  \cellcolor{red!19.6}New Democracy & 19.6\%\\
  \cellcolor{red!19.2}Patriotic Alliance & 19.2\%\\
  \cellcolor{red!19}The River & 19\%\\
  \cellcolor{red!19}Polpular Unity & 19\%\\
  \cellcolor{red!18.5}Movement for Change & 18.5\%\\
  \cellcolor{red!17.4}Panhellenic Socialist Movement & 17.4\%\\
  \cellcolor{red!17.2}Democratic Left & 17.2\%\\
  \cellcolor{red!15.3}Democratic Renewal &15.3\%\\
  \cellcolor{red!14}Extra Parliamentary & 14\%\\
  \cellcolor{red!13.3}Popular Orthodox Rally & 13.3\%\\\hline\hline
  \cellcolor{red!6.3}Opposition & 6.3\%\\
  \cellcolor{red!0.3}Parliament & 0.3\%\\\hline
\end{tabular}
\caption{Average \textsc{disgust} score per political party. The color intensity reflects the score.}\vspace{-10pt}
\end{table}

Studying language evolution can reflect changes in the political and social sphere \cite{Montariol2021}, changes whose importance increases when they regard language used to stigmatise people. \citet{Rose2007} presented 250 labels used to stigmatise people with medical illness in school. In this work, we (a) explore the frequency of some of these terms in the parliamentary records, and (b) utilise emotion classification to investigate the evolution of the negative context they may appear in over time. Although static word embeddings (yet with multiple spaces) can be used to capture semantic shift and word usage change \cite{levy2015,goldberg}, as well as contextual embeddings to detect global context shifts \cite{Kellert2022}, we argue that \textit{emotional} context shifts also apply and that emotion classifiers can unlock their use to the study of language evolution. 

We employed the words `handicapped' (ανάπηρος), `disability' (ειδικές ανάγκες) and `crazy' (τρελός),\footnote{Each term corresponds to a group of derivative terms, including for example inflected word forms.} retrieving sentences comprising them from the Greek parliament corpus. We then sliced our corpus as in \cite{goldberg}, focusing on three periods: from 1989 to 2000, from 2001 to 2010, and from 2011 to 2020. From each decade we sampled 100 sentences, each of which was scored with \xpalo regarding the \textsc{disgust} emotion, in order to report the average \textsc{disgust} score per decade. These are words that describe specific conditions, hence we hypothesise that \textit{use in a negative emotional context indicates stigmatised use} and we are looking for an increased score over time. Statistical significance of the differences is computed with bootstrapping.\footnote{We repeated the algorithm one thousand times.} 

%

\noindent \textbf{Control groups} were created with the words `bad' and `good', repeating the same study, as well as with words related to politics whose usage could also be linked to stigma. One group comprised `racism' and `illegal immigrant' while the other comprised the words `communism', `capitalism', `left' and `right'. The support of all the selected words is shared in the Appendix (Table~\ref{fig:ece_support}).\footnote{We disregarded low-support terms such as: `spastic' (σπαστικος), `psychopath' (ψυχοπαθής), `gay' (ομοφυλόφιλος), `fascism' (φασισμός), and `feminism' (φεμινισμός).}

\begin{figure}[t]
\centering
  \includegraphics[width=.5\textwidth]{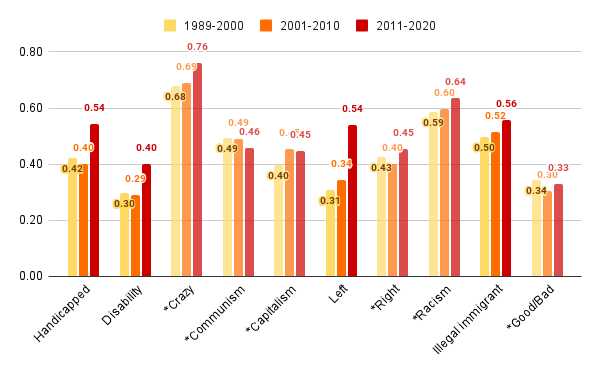}  \caption{\footnotesize Average \textsc{disgust} score computed on random samples per term (horizontally) per decade (in red the most recent). Faded colors (and stars aside the respective words) indicate a P-value that was greater than 0.05. }
  \label{fig:ece}\vspace{-10pt}
\end{figure}

\noindent \textbf{The results} are shown in Figure~\ref{fig:ece}. The words `handicapped' and `disability' show a statistically significant increase during the last decade. From our control groups, only the words `left' and `illegal immigrant' followed the same pattern.\footnote{Table \ref{table:p-values} in the Appendix presents all the P-values.}

\section{Conclusion}
This study presented three new datasets for emotion classification in Greek, which we release for public use. We benchmarked seven state of the art baselines that were based on machine, deep and transfer learning, and we showed that the performance for the emotion of `disgust' was comparable to that of sentiment and subjectivity detection. By using a model to score sentences from the Greek Parliament records, we detected months where the `disgust' emotion was generally high and we undertook a comparative analysis of the Greek political parties. Finally, we presented a way to use emotion classification in order to analyse the emotion evolution of a word's context, providing proof that two words describing a medical condition, are increasingly used in a negative emotional context in the Parliament, which may be alarming and should be further analysed in the future. 


\bibliography{anthology,custom}

\begin{thebibliography}{40}
\expandafter\ifx\csname natexlab\endcsname\relax\def\natexlab#1{#1}\fi

\bibitem[{Acheampong et~al.(2020)Acheampong, Wenyu, and
  Nunoo-Mensah}]{Acheampong2020}
Francisca~Adoma Acheampong, Chen Wenyu, and Henry Nunoo-Mensah. 2020.
\newblock Text-based emotion detection: Advances, challenges, and
  opportunities.
\newblock \emph{Engineering Reports}, 2(7):e12189.

\bibitem[{Ahmad et~al.(2020)Ahmad, Jindal, Ekbal, and
  Bhattachharyya}]{Ahmad2020}
Zishan Ahmad, Raghav Jindal, Asif Ekbal, and Pushpak Bhattachharyya. 2020.
\newblock Borrow from rich cousin: transfer learning for emotion detection
  using cross lingual embedding.
\newblock \emph{Expert Systems with Applications}, 139:112851.

\bibitem[{Alexandridis et~al.(2021)Alexandridis, Korovesis, Varlamis,
  Tsantilas, and Caridakis}]{Alexandridis2021}
Georgios Alexandridis, Konstantinos Korovesis, Iraklis Varlamis, Panagiotis
  Tsantilas, and George Caridakis. 2021.
\newblock Emotion detection on greek social media using bidirectional encoder
  representations from transformers.
\newblock In \emph{25th Pan-Hellenic Conference on Informatics}, pages 28--32.

\bibitem[{Athanasiou and Maragoudakis(2017)}]{athanasiou2017}
Vasileios Athanasiou and Manolis Maragoudakis. 2017.
\newblock \href {https://doi.org/10.3390/a10010034} {A novel, gradient boosting
  framework for sentiment analysis in languages where nlp resources are not
  plentiful: A case study for modern greek}.
\newblock \emph{Algorithms}, 10:34.

\bibitem[{Bagozzi et~al.(1999)Bagozzi, Gopinath, and Nyer}]{bagozzi1999}
Richard~P. Bagozzi, Mahesh Gopinath, and Prashanth~U. Nyer. 1999.
\newblock The role of emotions in marketing.
\newblock volume~27.

\bibitem[{Brave and Nass(2002)}]{brave2002}
Scott Brave and Clifford Nass. 2002.
\newblock \href {https://doi.org/10.1201/b10368-6} {Emotion in human–computer
  interaction}.
\newblock \emph{The Human-Computer Interaction Handbook: Fundamentals, Evolving
  Technologies and Emerging Applications}.

\bibitem[{Conneau et~al.(2019)Conneau, Khandelwal, Goyal, Chaudhary, Wenzek,
  Guzm{\'a}n, Grave, Ott, Zettlemoyer, and Stoyanov}]{conneau}
Alexis Conneau, Kartikay Khandelwal, Naman Goyal, Vishrav Chaudhary, Guillaume
  Wenzek, Francisco Guzm{\'a}n, Edouard Grave, Myle Ott, Luke Zettlemoyer, and
  Veselin Stoyanov. 2019.
\newblock Unsupervised cross-lingual representation learning at scale.
\newblock \emph{arXiv preprint arXiv:1911.02116}.

\bibitem[{Das et~al.(2021)Das, Sharif, Hoque, and Sarker}]{das2021}
Avishek Das, Omar Sharif, Mohammed~Moshiul Hoque, and Iqbal~H. Sarker. 2021.
\newblock \href {http://arxiv.org/abs/2104.08613} {Emotion classification in a
  resource constrained language using transformer-based approach}.

\bibitem[{Demszky et~al.(2020)Demszky, Movshovitz-Attias, Ko, Cowen, Nemade,
  and Ravi}]{Demszky2020}
Dorottya Demszky, Dana Movshovitz-Attias, Jeongwoo Ko, Alan Cowen, Gaurav
  Nemade, and Sujith Ravi. 2020.
\newblock \href {http://arxiv.org/abs/2005.00547} {Goemotions: A dataset of
  fine-grained emotions}.

\bibitem[{Desai et~al.(2020)Desai, Caragea, and Li}]{desai2020}
Shrey Desai, Cornelia Caragea, and Junyi~Jessy Li. 2020.
\newblock \href {http://arxiv.org/abs/2004.14299} {Detecting perceived emotions
  in hurricane disasters}.

\bibitem[{Devlin et~al.(2018)Devlin, Chang, Lee, and Toutanova}]{Devlin2018}
Jacob Devlin, Ming-Wei Chang, Kenton Lee, and Kristina Toutanova. 2018.
\newblock Bert: Pre-training of deep bidirectional transformers for language
  understanding.
\newblock \emph{arXiv preprint arXiv:1810.04805}.

\bibitem[{Fleiss and Cohen(1973)}]{fleiss}
J.~L. Fleiss and J.~Cohen. 1973.
\newblock \href {https://doi.org/10.1177/001316447303300309} {The equivalence
  of weighted kappa and the intraclass correlation coefficient as measures of
  reliability}.
\newblock In \emph{Educational and Psychological Measurement}, page 613–619,
  New Orleans, Louisiana.

\bibitem[{Gaind(2019)}]{Gaind2019}
Bharat Gaind. 2019.
\newblock Emotion detection and analysis on social media 1.

\bibitem[{Gonen et~al.(2020)Gonen, Jawahar, Seddah, and Goldberg}]{goldberg}
Hila Gonen, Ganesh Jawahar, Djam{\'e} Seddah, and Yoav Goldberg. 2020.
\newblock \href {https://doi.org/10.18653/v1/2020.acl-main.51} {Simple,
  interpretable and stable method for detecting words with usage change across
  corpora}.
\newblock In \emph{Proceedings of the 58th Annual Meeting of the Association
  for Computational Linguistics}, pages 538--555, Online. Association for
  Computational Linguistics.

\bibitem[{Hedderich et~al.(2020)Hedderich, Adelani, Zhu, Alabi, Markus, and
  Klakow}]{hedderich2020}
Michael~A. Hedderich, David Adelani, Dawei Zhu, Jesujoba Alabi, Udia Markus,
  and Dietrich Klakow. 2020.
\newblock \href {http://arxiv.org/abs/2010.03179} {Transfer learning and
  distant supervision for multilingual transformer models: A study on african
  languages}.

\bibitem[{Kabir and Madria(2021)}]{Kabir2021}
Md~Yasin Kabir and Sanjay Madria. 2021.
\newblock Emocov: Machine learning for emotion detection, analysis and
  visualization using covid-19 tweets.
\newblock \emph{Online Social Networks and Media}, 23:100135.

\bibitem[{Kant et~al.(2018)Kant, Puri, Yakovenko, and Catanzaro}]{kant2018}
Neel Kant, Raul Puri, Nikolai Yakovenko, and Bryan Catanzaro. 2018.
\newblock \href {http://arxiv.org/abs/1812.01207} {Practical text
  classification with large pre-trained language models}.

\bibitem[{Kellert and Zaman(2022)}]{Kellert2022}
Olga Kellert and Md~Mahmud~Uz Zaman. 2022.
\newblock Using neural topic models to track context shifts of words: a case
  study of covid-related terms before and after the lockdown in april 2020.
\newblock In \emph{Proceedings of the 3rd Workshop on Computational Approaches
  to Historical Language Change}, pages 131--139.

\bibitem[{Koutsikakis et~al.(2020)Koutsikakis, Chalkidis, Malakasiotis, and
  Androutsopoulos}]{koutsikakis2020}
John Koutsikakis, Ilias Chalkidis, Prodromos Malakasiotis, and Ion
  Androutsopoulos. 2020.
\newblock \href {https://doi.org/10.1145/3411408.3411440} {Greek-bert: The
  greeks visiting sesame street}.
\newblock \emph{11th Hellenic Conference on Artificial Intelligence}.

\bibitem[{Krommyda et~al.(2020)Krommyda, Rigos, Bouklas, and
  Amditis}]{Krommyda2020}
Maria Krommyda, Anastatios Rigos, Kostas Bouklas, and Angelos Amditis. 2020.
\newblock Emotion detection in twitter posts: a rule-based algorithm for
  annotated data acquisition.
\newblock In \emph{2020 International Conference on Computational Science and
  Computational Intelligence (CSCI)}, pages 257--262. IEEE.

\bibitem[{Kumar and Kumar(2021)}]{kumar2021}
Pedamuthevi~Kiran Kumar and Ishan Kumar. 2021.
\newblock Emotion detection and sentiment analysis of text.

\bibitem[{Lauscher et~al.(2020)Lauscher, Ravishankar, Vulić, and
  Glavaš}]{lauscher2020}
Anne Lauscher, Vinit Ravishankar, Ivan Vulić, and Goran Glavaš. 2020.
\newblock \href {http://arxiv.org/abs/2005.00633} {From zero to hero: On the
  limitations of zero-shot cross-lingual transfer with multilingual
  transformers}.

\bibitem[{Levy et~al.(2015)Levy, Goldberg, and Dagan}]{levy2015}
Omer Levy, Yoav Goldberg, and Ido Dagan. 2015.
\newblock \href {https://doi.org/10.1162/tacl_a_00134} {Improving
  distributional similarity with lessons learned from word embeddings}.
\newblock \emph{Transactions of the Association for Computational Linguistics},
  3:211--225.

\bibitem[{Markopoulos et~al.(2015)Markopoulos, Mikros, Iliadi, and
  Liontos}]{markopoulos2015}
George Markopoulos, George Mikros, Anastasia Iliadi, and Michalis Liontos.
  2015.
\newblock \href {https://doi.org/10.1007/978-3-319-15859-4_31} {Sentiment
  analysis of hotel reviews in greek: A comparison of unigram features}.
\newblock 9:373--383.

\bibitem[{Mohammad et~al.(2018)Mohammad, Bravo-Marquez, Salameh, and
  Kiritchenko}]{mohammad2018}
Saif Mohammad, Felipe Bravo-Marquez, Mohammad Salameh, and Svetlana
  Kiritchenko. 2018.
\newblock \href {https://doi.org/10.18653/v1/S18-1001} {{S}em{E}val-2018 task
  1: Affect in tweets}.
\newblock In \emph{Proceedings of The 12th International Workshop on Semantic
  Evaluation}, pages 1--17, New Orleans, Louisiana. Association for
  Computational Linguistics.

\bibitem[{Mohammad(2016)}]{Mohammad2016}
Saif~M Mohammad. 2016.
\newblock Sentiment analysis: Detecting valence, emotions, and other affectual
  states from text.
\newblock In \emph{Emotion measurement}, pages 201--237. Elsevier.

\bibitem[{Mohammad et~al.(2015)Mohammad, Zhu, Kiritchenko, and
  Martin}]{Mohammad2015}
Saif~M Mohammad, Xiaodan Zhu, Svetlana Kiritchenko, and Joel Martin. 2015.
\newblock Sentiment, emotion, purpose, and style in electoral tweets.
\newblock \emph{Information Processing \& Management}, 51(4):480--499.

\bibitem[{Montariol et~al.(2021)Montariol, Martinc, Pivovarova
  et~al.}]{Montariol2021}
Syrielle Montariol, Matej Martinc, Lidia Pivovarova, et~al. 2021.
\newblock Scalable and interpretable semantic change detection.
\newblock In \emph{Proceedings of the 2021 Conference of the North American
  Chapter of the Association for Computational Linguistics Human Language
  Technologies}. The Association for Computational Linguistics.

\bibitem[{Oberl{\"a}nder and Klinger(2018)}]{Oberlander2018}
Laura Ana~Maria Oberl{\"a}nder and Roman Klinger. 2018.
\newblock An analysis of annotated corpora for emotion classification in text.
\newblock In \emph{Proceedings of the 27th International Conference on
  Computational Linguistics}, pages 2104--2119.

\bibitem[{Palogiannidi et~al.(2016)Palogiannidi, Koutsakis, Losif, and
  Potamianos}]{Palogiannidi2016}
Elisavet Palogiannidi, Polychronis Koutsakis, E~Losif, and Alexandros
  Potamianos. 2016.
\newblock Affective lexicon creation for the greek language.

\bibitem[{Pires et~al.(2019)Pires, Schlinger, and Garrette}]{pires2019}
Telmo Pires, Eva Schlinger, and Dan Garrette. 2019.
\newblock \href {http://arxiv.org/abs/1906.01502} {How multilingual is
  multilingual bert?}

\bibitem[{Plutchik(1980)}]{plutchik}
Robert Plutchik. 1980.
\newblock A general psychoevolutionary theory of emotion.
\newblock In \emph{Theories of emotion}.

\bibitem[{Ranasinghe and Zampieri(2020)}]{ranasinghe2020}
Tharindu Ranasinghe and Marcos Zampieri. 2020.
\newblock \href {http://arxiv.org/abs/2010.05324} {Multilingual offensive
  language identification with cross-lingual embeddings}.

\bibitem[{Rose et~al.(2007)Rose, Thornicroft, Pinfold, and Kassam}]{Rose2007}
Diana Rose, Graham Thornicroft, Vanessa Pinfold, and Aliya Kassam. 2007.
\newblock 250 labels used to stigmatise people with mental illness.
\newblock \emph{BMC health services research}, 7(1):1--7.

\bibitem[{Rosenthal et~al.(2017)Rosenthal, Farra, and Nakov}]{semeval2017}
Sara Rosenthal, Noura Farra, and Preslav Nakov. 2017.
\newblock \href {https://doi.org/10.18653/v1/S17-2088} {{S}em{E}val-2017 task
  4: Sentiment analysis in {T}witter}.
\newblock In \emph{Proceedings of the 11th International Workshop on Semantic
  Evaluation ({S}em{E}val-2017)}, pages 502--518, Vancouver, Canada.
  Association for Computational Linguistics.

\bibitem[{Sailunaz et~al.(2018)Sailunaz, Dhaliwal, Rokne, and
  Alhajj}]{sailunaz2018}
Kashfia Sailunaz, Manmeet Dhaliwal, Jon Rokne, and Reda Alhajj. 2018.
\newblock Emotion detection from text and speech: a survey.
\newblock \emph{Social Network Analysis and Mining}, 8(1):1--26.

\bibitem[{Seyeditabari et~al.(2018)Seyeditabari, Tabari, and
  Zadrozny}]{Seyeditabari2018}
Armin Seyeditabari, Narges Tabari, and Wlodek Zadrozny. 2018.
\newblock Emotion detection in text: a review.
\newblock \emph{arXiv preprint arXiv:1806.00674}.

\bibitem[{Tela et~al.(2020)Tela, Woubie, and Hautamaki}]{tela2020}
Abrhalei Tela, Abraham Woubie, and Ville Hautamaki. 2020.
\newblock \href {http://arxiv.org/abs/2006.07698} {Transferring monolingual
  model to low-resource language: The case of tigrinya}.

\bibitem[{Vaswani et~al.(2017)Vaswani, Shazeer, Parmar, Uszkoreit, Jones,
  Gomez, Kaiser, and Polosukhin}]{vaswani2017}
Ashish Vaswani, Noam Shazeer, Niki Parmar, Jakob Uszkoreit, Llion Jones,
  Aidan~N. Gomez, Lukasz Kaiser, and Illia Polosukhin. 2017.
\newblock \href {http://arxiv.org/abs/1706.03762} {Attention is all you need}.

\bibitem[{Yang et~al.(2019)Yang, Dai, Yang, Carbonell, Salakhutdinov, and
  Le}]{xlnet}
Zhilin Yang, Zihang Dai, Yiming Yang, Jaime Carbonell, Russ~R Salakhutdinov,
  and Quoc~V Le. 2019.
\newblock Xlnet: Generalized autoregressive pretraining for language
  understanding.
\newblock \emph{Advances in neural information processing systems}, 32.

\end{thebibliography}
\bibliographystyle{acl_natbib}

\section*{Appendix}\label{sec:appendix}
\appendix{
\section{Annotation}\label{appendix:data}
The examples shown to the annotators of our dataset (\es and \palo), addressing the question: \textit{Which of the following options best describes the emotional state of the tweeter?}, are shown in Table~\ref{table:examples}. 

\begin{table}[ht]
\scalebox{0.65}{
\centering
\begin{tabular}{|l|} 
\hline
\begin{tabular}[c]{@{}l@{}}\textbf{anger (also includes annoyance, rage)}\\e.g~ \textit{“Εν τω μεταξύ όλοι δίνουν παράδειγμα την Παπαστράτος. }\\\textit{Τα ξενοδοχεία πως δουλεύουν ρε μπετόστοκοι; Είδατε Κυριακή }\\\textit{κλειστό ξενοδοχείο; Σκατά έχουν για μυαλό ρε μλκ τι να πω...}\\\textit{\#syriza\_xeftiles \#ΞΑΝΑΕΡΧΕΤΑΙ”}\end{tabular}                                                      \\ 
\hline
\begin{tabular}[c]{@{}l@{}}\textbf{anticipation (also includes interest, vigilance)}\\e.g :\textit{“Ελπίζω να καταφέρει να ανεβάσει ποιοτικά το νετφλιξ}\\\textit{ αν υπάρχει τέτοιο ενδεχόμενο”}\end{tabular}                                                                                                                                                                                             \\ 
\hline
\begin{tabular}[c]{@{}l@{}}\textbf{disgust (also includes disinterest, dislike, loathing)}\\e.g:\textit{“Παιδιά μια συμβουλή μακρυά από FORTHNET χαλαρα}\\\textit{ ότι από απαίσιο κυκλοφορεί σε Ίντερνετ”}\end{tabular}                                                                                                                                                                                   \\ 
\hline
\begin{tabular}[c]{@{}l@{}}\textbf{fear (also includes apprehension, anxiety, terror)}\\e.g:\textit{ “Φοβάμαι πως η επόμενη φάση της πανδημίας στη χώρα}\\\textit{ άρχισε νωρίτερα από ότι υπολογίζαμε. Το φθινόπωρο τα πράγματα}\\\textit{ είναι σχεδόν σίγουρο ότι θα εξελιχθούν σε ένα νέο (χειρότερο) }\\\textit{κύμα ή την διόγκωση του τωρινού, ακριβώς για τους λόγους που γράφεις.”}\end{tabular}  \\ 
\hline
\begin{tabular}[c]{@{}l@{}}\textbf{ joy (also includes serenity, ecstasy)}\\e.g:\textit{“Αυτός που μου δίνει τους κωδικούς πλήρωσε ΕΠΙΤΕΛΟΥΣ }\\\textit{το Νετφλιξ. Θα πάθω εγκεφαλικό απ τη χαρά μου.”}\end{tabular}                                                                                                                                                                                      \\ 
\hline
\begin{tabular}[c]{@{}l@{}}\textbf{sadness (also includes pensiveness, grief)}\\e.g\textit{:“Με λύπη μου σας λέω , ότι αν είστε συνδρομητής @COSMOTE}\\\textit{ κ έχετε τεχνική βλάβη , ούτε άκρη θα βρείτε Σάββατο Κυριακο κ }\\\textit{για την αποκατάσταση της, μπορεί να περιμένετε μια βδομάδα!!!!”}\end{tabular}                                                                                     \\ 
\hline
\begin{tabular}[c]{@{}l@{}}\textbf{surprise (also includes distraction, amazement)}\\e.g:\textit{“Υπέροχη νέα εφαρμογή Cosmote TV επιτέλους έχει και το Ε!”}\end{tabular}                                                                                                                                                                                                                                  \\ 
\hline
\begin{tabular}[c]{@{}l@{}}\textbf{trust (also includes acceptance, liking, admiration)}\\e.g:\textit{“@SpyrosLAP: Αυτό είναι πολύ καλό. Ώρα του Υπ Παιδείας }\\\textit{να πάρει τη χώρα μπροστά \#Cyprus \#Cyta @AnastasiadesCY}\\\textit{ \#ΜΕΝΟΥΜΕΣΠΙΤΙ \#StayAtHome”}\end{tabular}                                                                                                                     \\ 
\hline
\begin{tabular}[c]{@{}l@{}}\textbf{other (sarcasm,irony,or other emotion)}\\e.g:\textit{ “ΟΤΕ ακούς; Τηλεφωνώ στο 13888 από την Παρασκευή, }\\\textit{αλλά άκρα του τάφου σιωπή. Τί έπαθε ο γίγαντας των }\\\textit{τηλεπικοινωνιών μας; @COSMOTE”.}\end{tabular}                                                                                                                                          \\ 
\hline
\begin{tabular}[c]{@{}l@{}}\textbf{none}\\e.g:\textit{“Αυτές είναι οι νέες σειρές και οι νέες ταινίες που έρχονται }\\\textit{στο Netflix μέσα στο Δεκέμβριο! https://t.co/pxIpmDyZx1”}\end{tabular}                                                                                                                                                                                                       \\
\hline
\end{tabular}}
\caption{The options and the corresponding examples from the guidelines during the annotation for the development of our dataset.}
\label{table:examples}
\end{table}

The guidelines were updated with the note and the example of Table~\ref{tab:ext_guideline}, for the final annotation of \es and \palo parts.

\begin{table}[ht]
    \centering
    \begin{tabular}{|c|p{5cm}|}\hline
         \textsc{note} & \textit{If the tweet involves news/announcement, it should be classified in the \textit{‘none’} class, assuming that the author does not have the emotion expressed by the news} \\\hline
         \textsc{example} & ‘ΑΠΟΚΛΕΙΣΤΙΚΟ: Επίκαιρη Επερώτηση για NOVA και αθέμιτο ανταγωνισμό Μαρινάκη… καταθέτει ο ΣΥΡΙΖΑ! 'URL' μέσω του χρήστη @user’ ("EXCLUSIVE: Topical Question for NOVA and unfair competition Marinaki" SYRIZA testifies! 'URL' via @user')\\\hline 
    \end{tabular}
    \caption{Note and example added to the annotation guidelines during the development of the \es dataset.}
    \label{tab:ext_guideline}
\end{table}

\section{Experimental details}\label{appendix:exp}
GreekBERT and XLM-R were trained for 30 epochs with early stopping, patience of 3 epochs, batch size 16, learning rate 1e-5 for XLM-R and 5e-5 for GreekBERT, monitoring the validation loss, maximum length of 109 for XLM-R and 85 for GreekBERT.

\begin{table*}[ht]
\centering\tiny
\arrayrulecolor{black}
\begin{tabular}{l l l l l l l l l l l} 
\hline
& \multicolumn{9}{c}{\textbf{Emotion}} & \\ \hline
& \textbf{anger} & \bf antic. & \bf disgust & \textbf{fear} & \textbf{joy}  & sadness  & surprise & \textbf{trust} & \textbf{none} & \textbf{AVG}  \\\hline
\xzero & 0.38 (0.02) & 0.12 (0.01) & 0.82 (0.02) & 0.03 (0.00) & 0.49 (0.04) & 0.10 (0.02) & 0.07 (0.01) & 0.18 (0.03) & 0.92 (0.01) & 0.35 \\\hline
\xart & 0.33 (0.01) & 0.13 (0.01) & 0.68 (0.03) & 0.07 (0.01) & 0.31 (0.04) & 0.07 (0.01) & 0.05 (0.01) & 0.10 (0.01) & 0.89 (0.01) & 0.29 \\ \hline
\xartpalo & \textbf{0.51} (0.00)  & 0.43 (0.00) & 0.94 (0.00) & \textbf{0.15 (0.01)} & 0.50 (0.04)         & \textbf{0.19 (0.04)} & 0.06 (0.01) & 0.25 (0.01) & \textbf{0.99 (0.00)} & \textbf{0.45}  \\\hline
\xpalo      & 0.46 (0.01) & \textbf{0.50 (0.00)} & 0.93 (0.00) & 0.09 (0.01) & \textbf{0.54 (0.03)} & 0.04 (0.01) & \textbf{0.09 (0.02)} & \textbf{0.28 (0.02)}  & \textbf{0.99 (0.00)} & 0.44 \\\hline
\xnope & 0.43 (0.00) & 0.19 (0.01) & 0.90 (0.00) & 0.03 (0.01) & 0.48 (0.07) & 0.03 (0.01) & 0.03 (0.01) & 0.20 (0.13) & 0.98 (0.00) & 0.37 \\\hline
\gpalo & 0.49 (0.02) & 0.31 (0.09) & \textbf{0.95 (0.00)} & 0.03 (0.02) & 0.45 (0.09) & 0.03 (0.01) & 0.03 (0.01) & 0.24 (0.03) & 0.98 (0.00) & 0.39 \\\hline
\rfpalo & 0.34 (0.01)          & 0.14 (0.02) & 0.81 (0.01) & 0.05 (0.03) & 0.13 (0.02) & 0.02 (0.00) & 0.03 (0.01) & 0.10 (0.01) & 0.93 (0.00) & 0.28 \\\hline
\end{tabular}
\caption{AUPRC (average across three repetitions) of emotion classifiers with the standard error of the mean (SEM) in the brackets}
\label{table:emotion-sem}

\end{table*}

\begin{table*}[ht]
\centering
\begin{tabular}{
!{\color{black}\vrule}l!{\color{black}\vrule}l!{\color{black}\vrule}l!{\color{black}\vrule}l!{\color{black}\vrule}!{\color{black}\vrule}l!{\color{black}\vrule}!{\color{black}\vrule}l!{\color{black}\vrule}l!{\color{black}\vrule}!{\color{black}\vrule}l!{\color{black}\vrule}} 
\hline
& \multicolumn{3}{c}{\textbf{Sentiment}} & & \multicolumn{2}{c}{\textbf{Subjectivity}} & \\\hline
& \textbf{neg}  & \textbf{pos}  & \textbf{neu} & \textbf{AVG} & \textbf{subj}  & \textbf{obj} & \textbf{AVG}\\ 
\cline{1-7} \cline{8-8}
\xzero & 0.84 (0.01) & 0.40 (0.02) & 0.93 (0.01) & 0.72 & 0.80 (0.02) & 0.93 (0.01 & 0.86 \\ \hline
\xart & 0.69 (0.03) & 0.18 (0.03) & 0.90 (0.01) & 0.59 & 0.72 (0.03) & 0.90 (0.01) & 0.81 \\\hline
\xartpalo & 0.95 (0.00)          & 0.41 (0.00)         & \textbf{0.99} (0.00)                                      & 0.78          & \textbf{0.97} (0.00) & \textbf{0.99} (0.00)                                                         & \textbf{0.98}  \\ 
\hline
\xpalo      & 0.95 (0.00)          & \textbf{0.43} (0.02) & \textbf{0.99} (0.00)                                     & \textbf{0.79} & 0.96 (0.00)          & \textbf{0.99} (0.00)                                                        & \textbf{0.98}  \\ 
\hline
\xnope           & 0.93 (0.00)          & 0.39 (0.02)         & \textbf{0.99 (0.00)}                                      & 0.77 & 0.95 (0.01)           & \textbf{0.99} (0.01)                                                        & 0.97           \\ 
\hline

\gpalo         & \textbf{0.96} (0.00) & 0.39 (0.06)          & \textbf{0.99} (0.00)                                      & 0.78          & \textbf{0.97} (0.00)& \textbf{0.99} (0.00)                                                         & \textbf{0.98}  \\ 
\hline
\rfpalo               & 0.84 (0.01)         & 0.17 (0.01)         & 0.95 (0.00)                                               & 0.65          & 0.87 (0.01)         & 0.95 (0.00)                                                                  & 0.91           \\
\hline
\end{tabular}
\caption{AUPRC (average across three runs) of sentiment and subjectivity classifiers with the standard error of the mean (SEM) in the brackets.}
\label{table:sent and subj-sem}
\end{table*}
}

\section{Emotion detection in political speech}\label{appendix:emo_in_pol}

\subsection*{Model selection}
We manually evaluated our 3 best performing emotion detectors, vis. \xpalo, \gpalo, \xartpalo, on a small dataset, consisting of sentences that were randomly sampled from the Greek Parliament Corpus. \xpalo was found to perform slightly better in this sample, hence it was preferred over \xartpalo (one unit higher in AUPRC in \textsc{disgust}; see Table~\ref{table:emotion}) for this study.

\subsection*{Events potentially responsible for `disgust'}
Table~\ref{table:events} presents events that potentially rationalise the highest \textsc{disgust} scores in the respective months. These are September of 1991,\footnote{\url{https://www.newscenter.gr/politiki/970602/kontogiannopoylos-katalipseis-paideia}} April of 1992,\footnote{\url{https://en.wikipedia.org/wiki/Macedonia\_naming\_disput}} April of 1993,\footnote{\url{https://www.esiweb.org/macedonias-dispute-greece}} August of 1993,\footnote{\url{https://www.tovima.gr/2008/11/25/archive/pws-epese-o-mitsotakis/}} January of 2000,\footnote{\url{https://m.naftemporiki.gr/story/1844644/politikooikonomika-orosima-10-dekaetion}} March of 2000,\footnote{\url{https://en.wikipedia.org/wiki/2000\_Greek\_legislative\\\_election}} November of 2015,\footnote{\url{https://www.ertnews.gr/eidiseis/ellada/prosfigiki-krisi-ke-periferiakes-exelixis-sto-epikentro-tis-episkepsis-tsipra-stin-tourkia/}} April of 2015,\footnote{\url{https://www.theguardian.com/business/live/2015/apr/08/\\shell-makes-47bn-move-for-bg-group-live-updates}} January of 2019,\footnote{\url{https://www.euronews.com/2019/01/24/explained-the-controversial-name-dispute-between-greece-and-fyr-macedonia}} and May of 2019.\footnote{\url{}}

\begin{table}
\centering\small
\begin{tabular}{|p{1.5cm}|p{5.4cm}|} 
\hline
\textbf{Date}   & \textbf{Event}\\\hline
1991, Sep & Bill of the Minister of Education Vassilis Kontogiannopoulos brought reactions.\\\hline
1992, Apr & Meeting of political leaders; Macedonian issue.\\\hline
1993, Apr & FYROM officially becomes a member of the UN.\\\hline
1993, Aug & Quarrels leading to the fall of the government.\\\hline
2000, Jan & Finalization of the drachma exchange rate against the euro.\\\hline
2000, Mar & Elections New Democracy succeeds Panhellenic Socialist Movement.\\\hline
2015, Nov & The Greek Prime Minister visits the Turkish Prime Minister.\\\hline
2015, Apr & The Greek Prime Minister visits the Russian Prime Minister.\\\hline
2019, Jan & Macedonian Issue.\\\hline
2019, May & Loss in European elections leads to a call for early parliamentary elections.\\\hline
\end{tabular}
\caption{The months with the higher values of \textsc{disgust}, potentially rationalised by the shown events.}
\label{table:events}
\arrayrulecolor{black}
\end{table}

\begin{table*}[ht]
\centering
\begin{tabular}{|p{\textwidth}|}\hline
September 1991~ \\\hline
Πως, λοιπόν, να έχουμε εμπιστοσύνη ότι δε θα εφαρμοστούν και πάλι σε κάποια στιγμή μεσούσης-\\ για να θυμηθούμε την καθαρεύουσα- της σχολικής και εκπαιδευτικής χρονικής διάρκειας, \\κάποια μέτρα πάλι σαν και αυτά του κ. Κοντογιαννόπουλου που δεν έφεραν απλώς κρίση, έφεραν έκρηξη.  \\\hline
{April 1992} \\ \hline
Φτάσαμε στο γεγονός η κυβέρνηση της Βουλγαρίας και ο φίλος του κυρίου Πρωθυπουργού ο κ. Ζέλεφ\\ να αναγνωρίσει τα Σκόπια πριν καλά-καλά υπάρξουν.\\\hline
April 1993 \\ \hline
Νομίζω ότι αυτό στη σημερινή συγκυρία είναι απαράδεκτο, αν θέλουμε όλοι εμείς που φλυαρούμε για \\τη Μακεδονία να εννοούμε τελικά ότι εκεί υπάρχει ένα καινούργιο ζήτημα που πρέπει να \\αντιμετωπίσουμε με νέες προτεραιότητες και νέες ιεραρχήσεις\\ \hline
August 1993 \\ \hline
Για να δείτε, πόσο μακριά από την πραγματικότητα, ακόμη και σήμερα και όχι μόνο τα 8 χρόνια που\\ βρισκόσαστε στην κυβέρνηση, είσθε, ξεκομμένοι από την πραγματικότητα την ευρωπαϊκή, τη διεθνή\\ και παραπληροφορείτε τον Ελληνικό Λαό.\\ \hline
January 2000\\ \hline
Και νομίζω ότι αυτή η προκήρυξη τελικά οδήγησε στο να γίνει άλλη μια προσπάθεια διαρθρωτικής\\ αλλαγής στην οικονομία μας, τελείως ανεπιτυχής και να βάλει τη σφραγίδα στην Κυβέρνηση\\ της αποτυχίας, στην Κυβέρνηση που δεν έχει μέλλον τουλάχιστον για τη μετά ΟΝΕ εποχή \\\hline
March 2000\\\hline
Δηλαδή κάθε φορά θα κάνουμε εκλογές με άθλιες νομοθεσίες και θα υποσχόμαστε ότι μετά\\ τις εκλογές θα τα ξαναδούμε; Μα, το θέμα είναι με ποιους όρους διεξάγουμε τις εκλογές τώρα. \\\hline
November 2015 \\ \hline
Πήρε 3 δισεκατομμύρια σε ρευστό, πήρε βίζα για να εισέρχονται οι Τούρκοι και οι πάσης φύσεως\\ τζιχαντιστές ισλαμιστές στην Ευρωπαϊκή Ένωση και να κάνουν ό,τι θέλουν και άρχισαν και οι ενταξιακές\\ της διαπραγματεύσεις.\\\hline
April 2015 \\ \hline
Ακόμα και το φλερτ με τον Πούτιν και τη Ρωσία καταλήγει στο πουθενά.    \\\hline
January 2019 \\ \hline
Χέρι-χέρι ξεπουλάτε τη Μακεδονία μας, ΣΥΡΙΖΑ και Νέα Δημοκρατία. \\ \hline
May 2019 \\ 
\hline
Τι εννοώ δηλαδή: Επειδή τρομοκρατήθηκαν κάποιοι λεγόμενοι «κεντρώοι» ψηφοφόροι από τη συμπεριφορά\\ της ακροδεξιάς πτέρυγας της Νέας Δημοκρατίας, η οποία έχει επιβάλει τον νόμο της στην ηγεσία της Νέας\\ Δημοκρατίας, έρχεται τώρα η Νέα Δημοκρατία να δημιουργήσει ένα επικοινωνιακό αντίβαρο με βάση το ήθος\\ του Πολάκη και να συζητάμε πηγαίνοντας προς εκλογές για τον Πολάκη και όχι οποιοδήποτε θέμα είναι σοβαρό\\ και αφορά τη ζωή και την καθημερινότητα των πολιτών.  \\\hline
\end{tabular}
\caption{Parliamentary texts classified as \textsc{disgust}, selected from the 10 highest-scored months.}
\label{table:examples-disgust}
\end{table*}

\subsection*{Emotional context shift}

\begin{figure}[ht]
\centering
  \includegraphics[width=.45\textwidth]{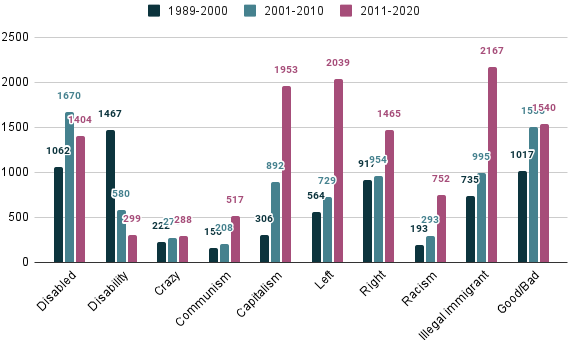}  \caption{\footnotesize Support of the target words per decade.}
  \label{fig:ece_support}
\end{figure}

The support of the selected terms is shown in Figure~\ref{fig:ece_support}, where we can see that the usage of half of them (i.e., `capitalism', `left', 'right', `racism', `illegal immigrant') is increased in the last decade.

The P-values of the target terms are shown in Table~\ref{table:p-values}. Besides the two words that are being used to stigmatise people with medical illness (top three rows), we observe that the context of the terms \textsc{left} and \textsc{illegal immigrant} is also being being shifted towards a more negative emotional state.

\begin{table}[ht!]\large
\centering
\begin{tabular}{|l|l|} 
\hline
\bf Target Term              & \bf P-value         \\\hline
\hline
\sc handicapped           & \textbf{0.027}  \\ 
\hline
\sc disability        & \textbf{0.026}  \\
\hline
\sc crazy             & 0.134           \\\hline  
\hline
\sc communism         & 0.328           \\ 
\hline
\sc capitalism        & 0.437           \\ 
\hline
\sc left              & \textbf{0.046}  \\ 
\hline
\sc right             & 0.404           \\ 
\hline
\sc racism            & 0.227           \\ 
\hline
\sc illegal immigrant & \textbf{0.019}  \\\hline 
\hline
\sc good/bad          & 0.209           \\
\hline
\end{tabular}
\caption{The target terms with their corresponding p-values. On the top are terms used to stigmatise people and lower are two control groups. In bold are values lower than 0.05.}
\label{table:p-values}
\end{table}

\end{document}